# Probabilistic Method of Measuring Linguistic Productivity

Sergei Monakhov, Friedrich Schiller University Jena (Germany)

sergei.monakhov@uni-jena.de

**Abstract**. In this paper I propose a new way of measuring affixes' productivity that objectively assesses the ability of an affix to be used to coin new complex words and, unlike other popular measures, is not directly dependent upon tokens' frequency. Specifically, I suggest that linguistic productivity may be viewed as the probability of an affix to combine with a random base. The advantages of this approach include the following: (1) token frequency does not dominate the productivity measure but naturally influences the sampling of bases; (2) we are not just counting attested word types with an affix but rather simulating the construction of these types and then checking whether they are attested in the corpus; and (3) a corpus-based approach and randomised design assure that true neologisms and words coined long ago have equal chances to be selected. The proposed algorithm is evaluated both on English and Russian data. The obtained results provide some valuable insights into the relation of linguistic productivity to the number of types and tokens. It looks like burgeoning linguistic productivity manifests itself in an increasing number of types. However, this process unfolds in two stages: first comes the increase in high-frequency items, and only then follows the increase in low-frequency items.

# 1 Introduction

The linguistic productivity of affixes has been an important topic of research for decades. It is most simply defined as '[t]he property of an affix to be used to coin new complex words' (Plag, 2018: 44). Plag, in his later work (2021), distinguishes between two possible approaches to linguistic productivity: it can be treated either categorically (qualitatively; cf. Bauer, 2001: 205) or continuously (quantitatively; cf. Bolinger, 1948: 18). Of these two approaches, as Plag remarks, the former is now mostly abandoned, and the latter is preferred. In what follows, I will adhere to the frequentist idea that totally unproductive and fully productive processes are end-points on a continuous scale, with infinitely many intermediate stages in between.

With this idea in mind, one needs a reliable way of measuring how productive a specific affix is. Many measures have been proposed in the literature so far; for example:

(i) the number of attested types (i.e., different words) with a given affix at a given point in time;

(ii) the ratio of the number of attested words with a given affix to the number of words that could, in principle, be formed with that affix (Aronoff, 1976);

(iii) the number of neologisms with a given affix at a given point in time (Plag, 2021);

(iv) Baayen's set of measures (Baayen and Lieber, 1991; Baayen 1992, 1993, 1994, 2009; Baayen and Renouf, 1996):

(iv.i) 'expanding productivity' (or 'hapax-conditioned degree of productivity')— the ratio of the number of hapax legomena with a given affix to the total number of hapax legomena in a given corpus;

(iv.ii) 'potential productivity' (or 'category-conditioned degree of productivity')— the ratio of the number of hapax legomena with a given affix to the total number of tokens with that affix in a given corpus.

Of all these different measures, the one that has become the most well-known and widespread in the literature is the so-called 'potential productivity', which can be interpreted as follows: 'a large number of hapaxes lead to a high value of P, thus indicating a productive morphological process. Conversely, large numbers of high-frequency items lead <…> to a decrease of P, indicating low productivity' (Plag, 2021: 488). This measure has been used and continues to be referred to in multiple studies until the present day (Fernández-Domínguez, Díaz-Negrillo, and Štekauer, 2007; Plag and Baayen, 2009; Zirkel, 2010; Marzi and Ferro, 2014; Mendaza, 2015; Pierrehumbert and Granell, 2018, among others).

All this time, the idea of a 'category-conditioned degree of productivity' and the theories that grew out of it have not been seriously challenged (see, however, Bauer, 2001; Gaeta and Ricca, 2006; Pustylnikov and Schneider-Wiejowski, 2010). Gaeta and Ricca (2006) pointed out that the measure is ill-suited for the comparison of affixes with different token numbers since one will always overestimate the values of productivity for the less-frequent constructions. Unfortunately, the improvement the authors proposed (to compare the counts of hapaxes when equal numbers of tokens have been sampled for each affix) does not change the overall picture (Baayen, 2009: 905).

As I see it, the main problem with the hapax-based measure, besides its dependance upon token frequency, is that calculating the productivity measures of different affixes is not a matter of linguistic bookkeeping. Rather, we are interested in these measures because we want to understand how morphological productivity works (i.e. how derivational patterns spread). Of special importance here is the question of how the type and token frequency of linguistic items contribute to derivational patterns' self-propagation. From this standpoint, having a measure of linguistic productivity heavily influenced by the number of tokens is a somewhat unwelcome premise.

In what follows I would like to suggest a way of measuring affixes' productivity that objectively assesses '[t]he property of an affix to be used to coin new complex words' (Plag, 2018:

44) and is not directly dependent upon tokens' frequency. Specifically, I suggest that linguistic productivity may be viewed as the probability of an affix to combine with a random base. The advantages of this approach include the following: (1) token frequency does not dominate the productivity measure but naturally influences the sampling of bases; (2) we are not just counting attested word types with an affix but rather simulating the construction of these types and then checking whether they are attested in the corpus; and (3) a corpus-based approach and randomised design assure that true neologisms and words coined long ago have equal chances to be selected.

Currently, the procedure has only been tested on prefixes, but its basic principles readily extend to suffixes.

## 2 Algorithm for measuring linguistic productivity

The process of obtaining a linguistic productivity measure for a specific prefix consists of two parts. The first part runs as follows:

1. For each prefix $P_i$:

   1.1. A random sample of 100 content words $V = \{V_1, \ldots, V_{100}\}$ is drawn from a corpus of modern language.

      1.1.1. Each word from the sample is automatically checked for whether it contains any prefix, and if so, it is stripped of it, which results in the set of bases $B = \{B_1, \ldots, B_{100}\}$.

      1.1.2. The prefix $P_i$ is automatically combined with each base $B_j \in B$, which results in the set of $P_i$-prefixed words $PB = \{PB_1, \ldots, PB_{100}\}$

      1.1.3. The frequency of each $P_i$-prefixed word $PB_j \in PB$ is checked in the same corpus and recorded, which results in the set of values $F = \{F_1, \ldots, F_{100}\}$.

Upon completing the first part of the algorithm, one could get a naive estimation of the linguistic productivity measure that is based on the observed data. For example, one could obtain from F a subset of values FNZ, such that each value $FNZ_j \in F$, $FNZ_j \neq 0$, $j \leq 100$, and then calculate productivity of the prefix $P_i$ as simple relative frequency n(FNZ) / n(F). However, we are interested in the probabilistic assessment of linguistic productivity. One way to estimate it is to try to predict what this value will be for the 101st base; that is, for the first base coming out of the sample. This constitutes the second part of the algorithm.

To implement it, I constructed a dynamic Bayesian network consisting of (1) a two-node DAG, where node $X_0$ represents an observation of a prefix-base combination at time $t$, and node $X_1$ represents an observation of a prefix-base combination at time $t + 1$, as well as (2) a joint probability distribution of the following form (Neapolitan and Jiang, 2007):

$$P_0(\mathrm{x}[0]) \prod_{t=0}^{T-1} P_{\rightarrow}(\mathrm{x}[t+1]\mathrm{x}[t])$$

The cardinality of each node is equal to three, where 0 stands for no occurrence of a particular prefix-base combination in the corpus (indicating that the prefix does not combine with this base); 1 stands for a low-frequency occurrence (a prefix-base combination was considered of low frequency if its number of hits in the corpus was lower than the 0.5 quantile of the previously obtained results in the sample); and 2 stands for a high-frequency occurrence (number of hits in the corpus greater than or equal to the 0.5 quantile of the previously obtained results).

The first part of the joint probability distribution, representing the initial state, was parametrised as follows: $P(X_0 = 0) = 0.4$, $P(X_0 = 1) = 0.4$, $P(X_0 = 2) = 0.2$. The second part, which contains transition probabilities for estimating unknown states given some known observations, was parametrised as follows (Table 1):

Table 1. Prior transition probabilities for the dynamic Bayesian network

|  | $X_1 = 0$ | $X_1 = 1$ | $X_1 = 2$ |
|---|---|---|---|
| $X_0 = 0$ | 0.7 | 0.2 | 0.1 |
| $X_0 = 1$ | 0.2 | 0.7 | 0.1 |
| $X_0 = 2$ | 0.1 | 0.2 | 0.7 |

These probabilities mirror my prior beliefs about the distribution of lexical items; however, they were used only once to predict the outcome of the second prefix-base combination in the sample. At each subsequent step, both initial and transition probabilities were updated based on the observed evidence. Thus, the value of $P(X_1[t+1] = 1) + P(X_1[t+1] = 2)$ (or, equivalently, $1 - P(X_1[t+1] = 0)$) evaluated at time $t = 100$ constitutes the true value of linguistic productivity.

## 3 Probabilistic productivity measure and its insights

Using the algorithm described above and the English internet corpus from 2018 provided by Sketch Engine (*ententen18_tt31*; 21,926,740,748 words), I obtained productivity values for 25 English prefixes: *em-*, *en-*, *im-*, *trans-*, *con-*, *dis-*, *fore-*, *mis-*, *de-*, *counter-*, *inter-*, *cross-*, *mid-*, *under-*, *anti-*, *self-*, *out-*, *super-*, *over-*, *sub-*, *un-*, *re-*, *non-*, *pre-*, *in-*. Hyphenated and non-hyphenated variants were calculated separately and then summed up. It is instructive to compare, for all prefixes, the dynamics of the probabilities $P(X_1[t+1] = 0 \mid X_0[t])$ (blue lines of all subplots in Figure 1), $P(X_1[t+1] = 1 \mid X_0[t])$ (orange lines), and $P(X_1[t+1] = 2 \mid X_0[t])$ (green lines), for $t \in \{1, \ldots, 100\}$.

First, it is clear that all three probabilities, after initial uncertainty, converge towards the end of the range of sampling to some stationary distribution, as the variance of their values approaches zero. Thus, the aforementioned way of calculating linguistic productivity as the probability of a given prefix to combine with the first base outside of a sample of 100 random bases seems justified.

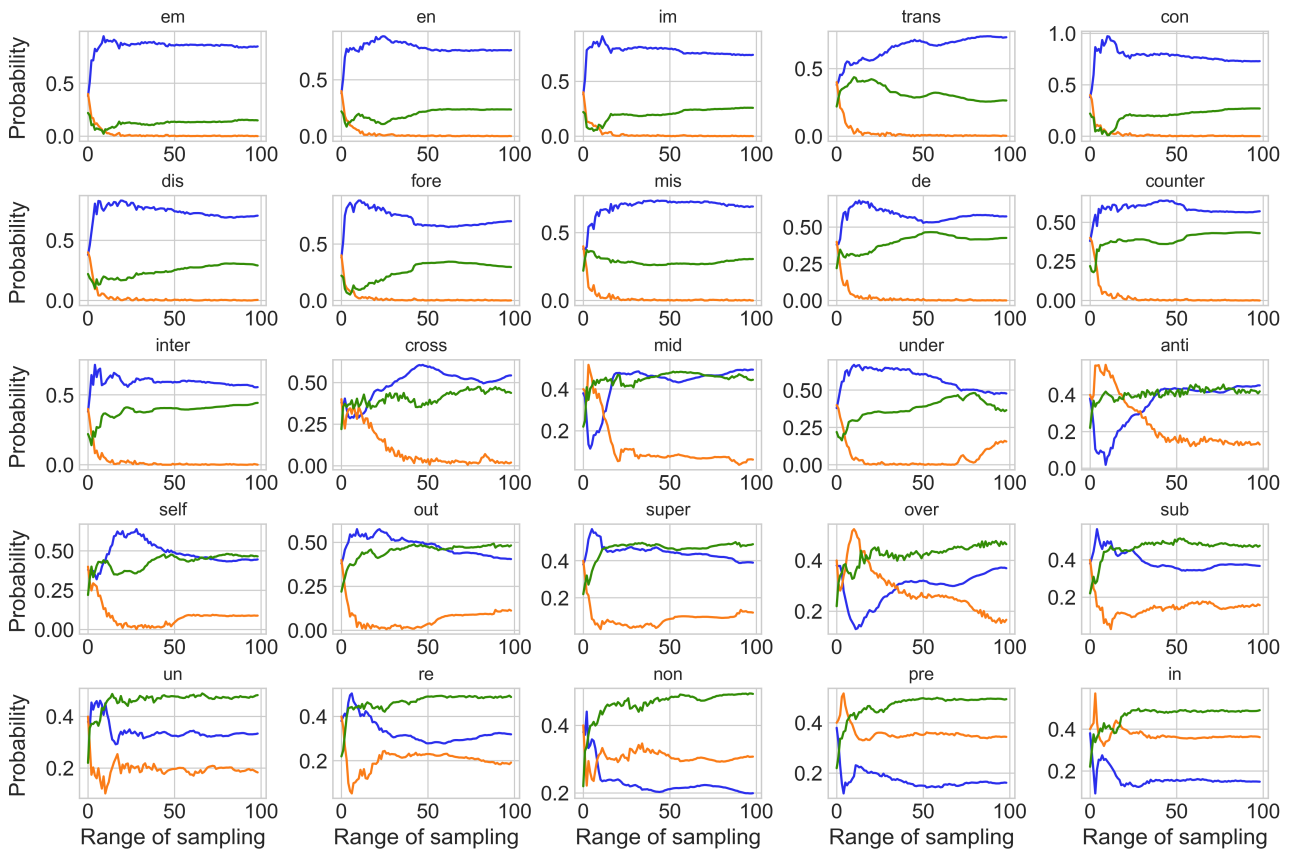

Figure 1. Probabilities of different values of X₁[t+1] evaluated at times $t \in \{1, …, 100\}$ for 25 English prefixes

Second, if one takes into account that the prefixes in Figure 1 are arranged, left-to-right, top-to-bottom, in order of ascending productivity measure, one interesting phenomenon becomes evident that sheds some new light on the long-standing debate on what is more important for linguistic productivity: the number of types or number of tokens. Concerning the final evaluations of the probabilities, all prefixes can be subdivided into three large groups, depending on the hierarchical order of these evaluations (Table 2). The first group encompasses prefixes from *em-* to *anti-*, and the probabilities here are arranged in the following hierarchy: $P(X_1[t+1] = 0 \mid X_0[t]) > P(X_1[t+1] = 2 \mid X_0[t]) > P(X_1[t+1] = 1 \mid X_0[t])$. In the second group, one finds prefixes from *self-* to *re-*, with the probabilities arranged in this way: $P(X_1[t+1] = 2 \mid X_0[t]) > P(X_1[t+1] = 0 \mid X_0[t]) > P(X_1[t+1] = 1 \mid X_0[t])$. Finally, the prefixes *non-*, *pre-*, and *in-* belong to the last group: $P(X_1[t+1] = 2 \mid X_0[t]) > P(X_1[t+1] = 1 \mid X_0[t]) > P(X_1[t+1] = 0 \mid X_0[t])$.

What is more interesting is that these categorical differences emerge as manifestations of an inherently gradient structure. From Table 2, where the prefixes are arranged in order of ascending productivity, it can be seen that for the first group (*0_2_1*), the differences in probabilities $P(X_1[t+1] = 0 \mid X_0[t])$ and $P(X_1[t+1] = 2 \mid X_0[t])$ continuously decrease, while the differences in probabilities $P(X_1[t+1] = 2 \mid X_0[t])$ and $P(X_1[t+1] = 1 \mid X_0[t])$ continuously increase (with some minor fluctuations). Visually, in the succession of subplots in Figure 1, this process can be described in terms of a green curve climbing higher and higher, with the other two curves held constant until finally, it changes places with a blue one, thus opening up the second group of prefixes.

For this second group (*2_0_1*), a similar mechanism of change can be observed, though with different contrasts. The differences in probabilities $P(X_1[t+1] = 2 \mid X_0[t])$ and $P(X_1[t+1] = 0 \mid X_0[t])$ become bigger, while the differences in probabilities $P(X_1[t+1] = 0 \mid X_0[t])$ and $P(X_1[t+1] = 1 \mid X_0[t])$ become smaller. Again, across the respective subplots of Figure 1, this process can be roughly described as that of a blue curve falling down and swapping near the bottom with an orange one.

The third group (*2_1_0*), though smallest, is of the same gradient nature. The gap between probabilities $P(X_1[t+1] = 2 \mid X_0[t])$ and $P(X_1[t+1] = 1 \mid X_0[t])$ successively narrows, while the gap between probabilities $P(X_1[t+1] = 1 \mid X_0[t])$ and $P(X_1[t+1] = 0 \mid X_0[t])$ widens. For simplicity, one can visualise an orange curve in Figure 1 approaching a green one at the top of the plot.

Table 2. English prefixes' productivity measures arranged in ascending order and divided into

groups

| group | prefix | product. | X_1 = 0 | X_1 = 1 | X_1 = 2 | contr._1 | diff._1 | contr._2 | diff._2 |
|-------|--------|----------|---------|---------|---------|----------|---------|----------|---------|
| 0_2_1 | *em-* | 0.150 | 0.850 | 0.001 | 0.150 | 0_2 | 0.700 | 2_1 | 0.149 |
| 0_2_1 | *en-* | 0.237 | 0.763 | 0.001 | 0.236 | 0_2 | 0.526 | 2_1 | 0.236 |
| 0_2_1 | *im-* | 0.259 | 0.741 | 0.000 | 0.258 | 0_2 | 0.483 | 2_1 | 0.258 |
| 0_2_1 | *trans-* | 0.268 | 0.732 | 0.004 | 0.264 | 0_2 | 0.467 | 2_1 | 0.261 |
| 0_2_1 | *con-* | 0.270 | 0.730 | 0.001 | 0.269 | 0_2 | 0.461 | 2_1 | 0.269 |
| 0_2_1 | *dis-* | 0.296 | 0.704 | 0.005 | 0.291 | 0_2 | 0.413 | 2_1 | 0.286 |
| 0_2_1 | *fore-* | 0.298 | 0.702 | 0.001 | 0.297 | 0_2 | 0.406 | 2_1 | 0.296 |
| 0_2_1 | *mis-* | 0.307 | 0.693 | 0.000 | 0.306 | 0_2 | 0.387 | 2_1 | 0.306 |
| 0_2_1 | *de-* | 0.427 | 0.573 | 0.001 | 0.426 | 0_2 | 0.147 | 2_1 | 0.426 |
| 0_2_1 | *counter-* | 0.430 | 0.570 | 0.000 | 0.430 | 0_2 | 0.140 | 2_1 | 0.430 |
| 0_2_1 | *inter-* | 0.444 | 0.556 | 0.001 | 0.443 | 0_2 | 0.113 | 2_1 | 0.442 |
| 0_2_1 | *cross-* | 0.457 | 0.543 | 0.019 | 0.438 | 0_2 | 0.105 | 2_1 | 0.419 |
| 0_2_1 | *mid-* | 0.507 | 0.493 | 0.062 | 0.445 | 0_2 | 0.048 | 2_1 | 0.383 |
| 0_2_1 | *under-* | 0.522 | 0.478 | 0.156 | 0.366 | 0_2 | 0.112 | 2_1 | 0.210 |
| 0_2_1 | *anti-* | 0.548 | 0.452 | 0.129 | 0.419 | 0_2 | 0.033 | 2_1 | 0.290 |
| 2_0_1 | *self-* | 0.554 | 0.446 | 0.089 | 0.465 | 2_0 | 0.019 | 0_1 | 0.357 |
| 2_0_1 | *out-* | 0.595 | 0.405 | 0.112 | 0.483 | 2_0 | 0.078 | 0_1 | 0.293 |
| 2_0_1 | *super-* | 0.610 | 0.390 | 0.121 | 0.489 | 2_0 | 0.099 | 0_1 | 0.269 |
| 2_0_1 | *over-* | 0.631 | 0.369 | 0.167 | 0.464 | 2_0 | 0.095 | 0_1 | 0.202 |
| 2_0_1 | *sub-* | 0.633 | 0.367 | 0.156 | 0.476 | 2_0 | 0.109 | 0_1 | 0.211 |
| 2_0_1 | *un-* | 0.666 | 0.334 | 0.183 | 0.483 | 2_0 | 0.149 | 0_1 | 0.151 |
| 2_0_1 | *re-* | 0.680 | 0.320 | 0.192 | 0.488 | 2_0 | 0.169 | 0_1 | 0.127 |
| 2_1_0 | *non-* | 0.801 | 0.199 | 0.308 | 0.493 | 2_1 | 0.185 | 1_0 | 0.108 |
| 2_1_0 | *pre-* | 0.838 | 0.162 | 0.344 | 0.494 | 2_1 | 0.150 | 1_0 | 0.182 |
| 2_1_0 | *in-* | 0.851 | 0.149 | 0.361 | 0.490 | 2_1 | 0.129 | 1_0 | 0.213 |

To understand what all of the above tells us about the relation of linguistic productivity to the number of types and tokens, we need to recall exactly what each type of probabilistic hierarchy signifies. The arrangement of probabilities for each group may be interpreted as follows. Group *0_2_1*: a small number of types with a few occasional high-frequency tokens. Group *2_0_1*: a substantial number of high-frequency tokens but a very limited number of types overall. Group *2_1_0*: many types and many tokens. The gradient nature of the transitions from group to group provides some evidence as to how the linguistic productivity of affixes grows. Burgeoning linguistic productivity manifests itself in an increasing number of types. However, this process unfolds in two stages: first comes the increase in high-frequency items, and only then follows the increase in low-frequency items.

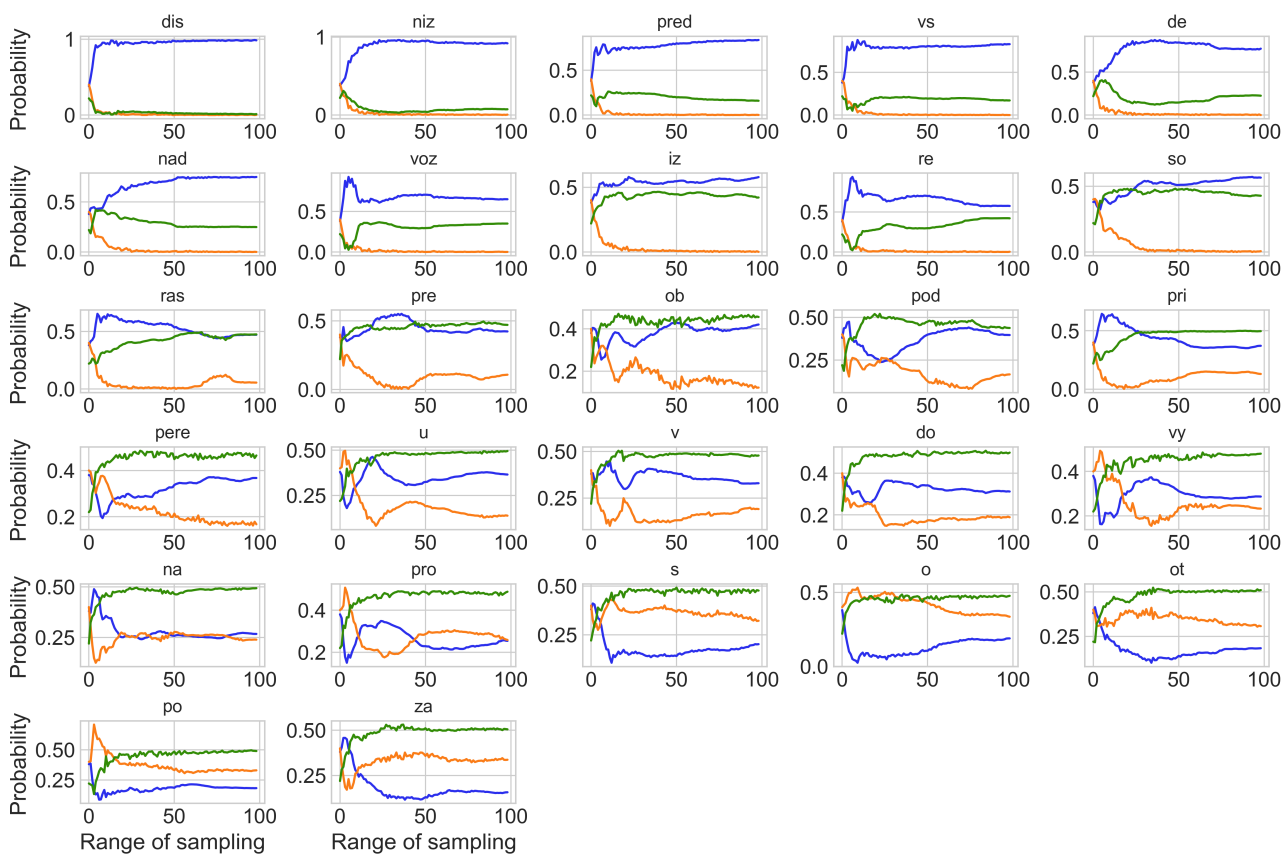

Figure 2. Probabilities of different values of $X_1[t+1]$ evaluated at times $t \in \{1, \ldots, 100\}$ for 27 Russian prefixes

In order to make sure that these patterns are not language-specific, I used the Russian internet corpus from 2011 provided by Sketch Engine (*rutenten11_8*; 14,553,856,113 words) to repeat the whole procedure of measuring linguistic productivity with 27 Russian verbal prefixes: *dis-*, *niz-*, *pred-*, *vs-*, *de-*, *nad-*, *voz-*, *iz-*, *re-*, *so-*, *ras-*, *pre-*, *ob-*, *pod-*, *pri-*, *pere-*, *u-*, *v-*, *do-*, *vy-*, *na-*, *pro-*, *s-*, *o-*, *ot-*, *po-*, *za-*. The results are provided in Figure 2 and Table 3. The similarity between the two languages is quite remarkable: one can easily identify in the Russian data the same tripartite division of affixes with the same fuzzy boundaries between the groups as in English.

Table 3. Russian prefixes' productivity measures arranged in ascending order and divided into groups

| group | prefix | product. | $X_1 = 0$ | $X_1 = 1$ | $X_1 = 2$ | contr._1 | diff._1 | contr._2 | diff._2 |
|-------|--------|----------|-----------|-----------|-----------|----------|---------|----------|---------|
| 0_2_1 | *dis-*  | 0.015 | 0.985 | 0.002 | 0.013 | 0_2 | 0.972 | 2_1 | 0.011 |
| 0_2_1 | *niz-*  | 0.080 | 0.920 | 0.006 | 0.074 | 0_2 | 0.845 | 2_1 | 0.069 |
| 0_2_1 | *pred-* | 0.161 | 0.839 | 0.001 | 0.160 | 0_2 | 0.679 | 2_1 | 0.158 |
| 0_2_1 | *vs-*   | 0.171 | 0.829 | 0.000 | 0.171 | 0_2 | 0.658 | 2_1 | 0.171 |
| 0_2_1 | *de-*   | 0.227 | 0.773 | 0.001 | 0.226 | 0_2 | 0.547 | 2_1 | 0.225 |
| 0_2_1 | *nad-*  | 0.250 | 0.750 | 0.001 | 0.249 | 0_2 | 0.501 | 2_1 | 0.249 |
| 0_2_1 | *voz-*  | 0.350 | 0.650 | 0.001 | 0.350 | 0_2 | 0.300 | 2_1 | 0.349 |
| 0_2_1 | *iz-*   | 0.422 | 0.578 | 0.001 | 0.421 | 0_2 | 0.157 | 2_1 | 0.420 |
| 0_2_1 | *re-*   | 0.424 | 0.576 | 0.003 | 0.421 | 0_2 | 0.155 | 2_1 | 0.419 |
| 0_2_1 | *so-*   | 0.435 | 0.565 | 0.007 | 0.428 | 0_2 | 0.137 | 2_1 | 0.421 |
| 2_0_1 | *ras-*  | 0.529 | 0.471 | 0.055 | 0.474 | 2_0 | 0.003 | 0_1 | 0.416 |
| 2_0_1 | *pre-*  | 0.577 | 0.423 | 0.107 | 0.470 | 2_0 | 0.048 | 0_1 | 0.315 |
| 2_0_1 | *ob-*   | 0.579 | 0.421 | 0.123 | 0.456 | 2_0 | 0.036 | 0_1 | 0.297 |
| 2_0_1 | *pod-*  | 0.604 | 0.396 | 0.166 | 0.438 | 2_0 | 0.041 | 0_1 | 0.231 |
| 2_0_1 | *pri-*  | 0.628 | 0.372 | 0.131 | 0.497 | 2_0 | 0.125 | 0_1 | 0.241 |
| 2_0_1 | *pere-* | 0.633 | 0.367 | 0.168 | 0.465 | 2_0 | 0.097 | 0_1 | 0.200 |
| 2_0_1 | u- | 0.634 | 0.366 | 0.139 | 0.495 | 2_0 | 0.129 | 0_1 | 0.227 |
| 2_0_1 | v- | 0.670 | 0.330 | 0.192 | 0.478 | 2_0 | 0.148 | 0_1 | 0.138 |
| 2_0_1 | do- | 0.688 | 0.312 | 0.189 | 0.499 | 2_0 | 0.187 | 0_1 | 0.124 |
| 2_0_1 | vy- | 0.713 | 0.287 | 0.233 | 0.480 | 2_0 | 0.193 | 0_1 | 0.054 |
| 2_0_1 | na- | 0.732 | 0.268 | 0.239 | 0.493 | 2_0 | 0.225 | 0_1 | 0.028 |
| 2_1_0 | pro- | 0.746 | 0.254 | 0.258 | 0.487 | 2_1 | 0.229 | 1_0 | 0.004 |
| 2_1_0 | s- | 0.799 | 0.201 | 0.322 | 0.477 | 2_1 | 0.155 | 1_0 | 0.120 |
| 2_1_0 | o- | 0.811 | 0.189 | 0.335 | 0.476 | 2_1 | 0.141 | 1_0 | 0.147 |
| 2_1_0 | ot- | 0.817 | 0.183 | 0.307 | 0.510 | 2_1 | 0.203 | 1_0 | 0.124 |
| 2_1_0 | po- | 0.818 | 0.182 | 0.329 | 0.489 | 2_1 | 0.160 | 1_0 | 0.148 |
| 2_1_0 | za- | 0.842 | 0.158 | 0.337 | 0.505 | 2_1 | 0.168 | 1_0 | 0.179 |

## 4 Comparing probabilistic and hapax-based productivity measures

It is instructive to compare my English results (blue line in Figure 3) plotted against hapax-based productivity measures obtained by Hay and Baayen (2002) for the same set of prefixes (orange line in Figure 3; original values were multiplied by 10 for comparability). Substantial differences between these two types of assessment are observable, and some inconsistencies are striking. I am doubtful, for example, how justified it is to place *non-*, with a hapax-based measure, at a so much higher productivity level compared to *un-* (0.07 vs. 0.005, i.e. 14 times greater). My data show that *un-* is extremely productive in modern English; it can combine with almost any base, whether verbal or nominal; to name some examples: (i) *It's such an **un-Michigan** thing to do*; (ii) *Fresh off the trail-blazing heels of SXSW comes Skillshare, the '**un-university**' of online universities*; (iii) *But you can't **un-shoot** a person*.

The top status of *in-*, based on my measure, may look surprising. However, one should take into account that there are two homonymous prefixes: a negative *in-*, as in *inaccurate*, and a prepositional *in-*, as in the following examples: (i) *This precludes **in-contact** operation*; (ii) *<…> this was a great **inlook** at movie industry*; (iii) *<…> **inmouth** or eye exposure occurs <…>*;

(iv) *Practice the wrong technique and it will be locked in and be hard to **in-do***; (v) *My favorite **in-major** class was Software Engineering.* While the former *in-* is only of very limited productivity in modern English, the latter one knows almost no bounds. Unfortunately, right now, it seems impossible for the proposed algorithm to distinguish between the two, so formal equivalence inevitably leads to overlap in the results.

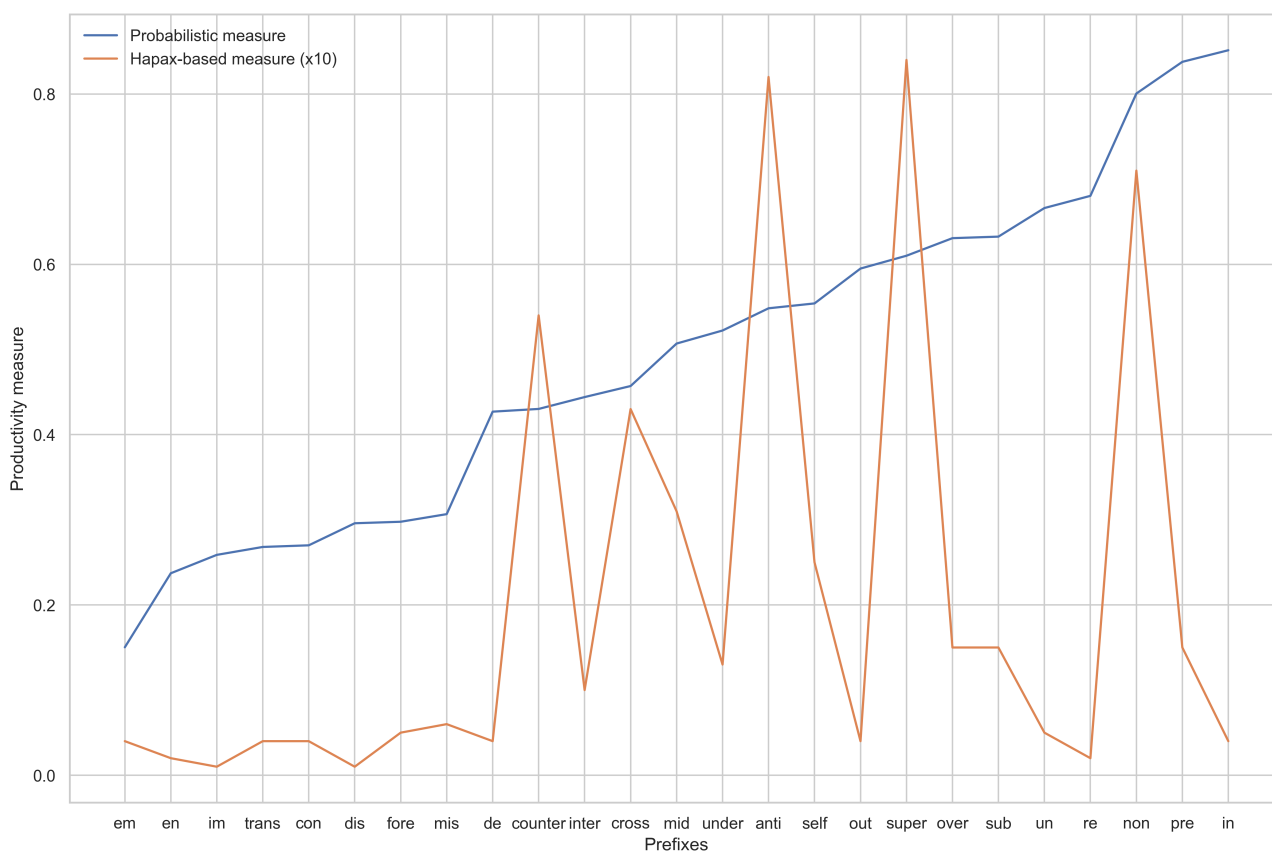

Figure 3. English prefixes' productivity measures calculated as probabilities of combining with a random base and as ratios of the number of hapaxes to the number of tokens

To better understand the nature of the difference between the two measures of linguistic productivity, I plotted both against the lines showing the number of types and number of tokens for each prefix (Figure 4). These values were obtained from the random samples of 100 bases that I used to train my dynamic Bayesian network. The results are conspicuous. The hapax-based measure is insignificantly correlated both with the number of types ($r = 0.35$, $p = 0.08$) and with the sum of tokens ($r = -0.12$, $p = 0.54$). Probabilistic measure, on the other hand, is not only perfectly correlated with the number of types but also significantly positively correlated with the sum of tokens ($r = 0.54$, $p < 0.01$). Noteworthy, a very similar positive correlation between prefixes' probabilistic productivity values and these prefixes' total sums of tokens was observed with the Russian data: $r = 0.57$, $p < 0.001$.

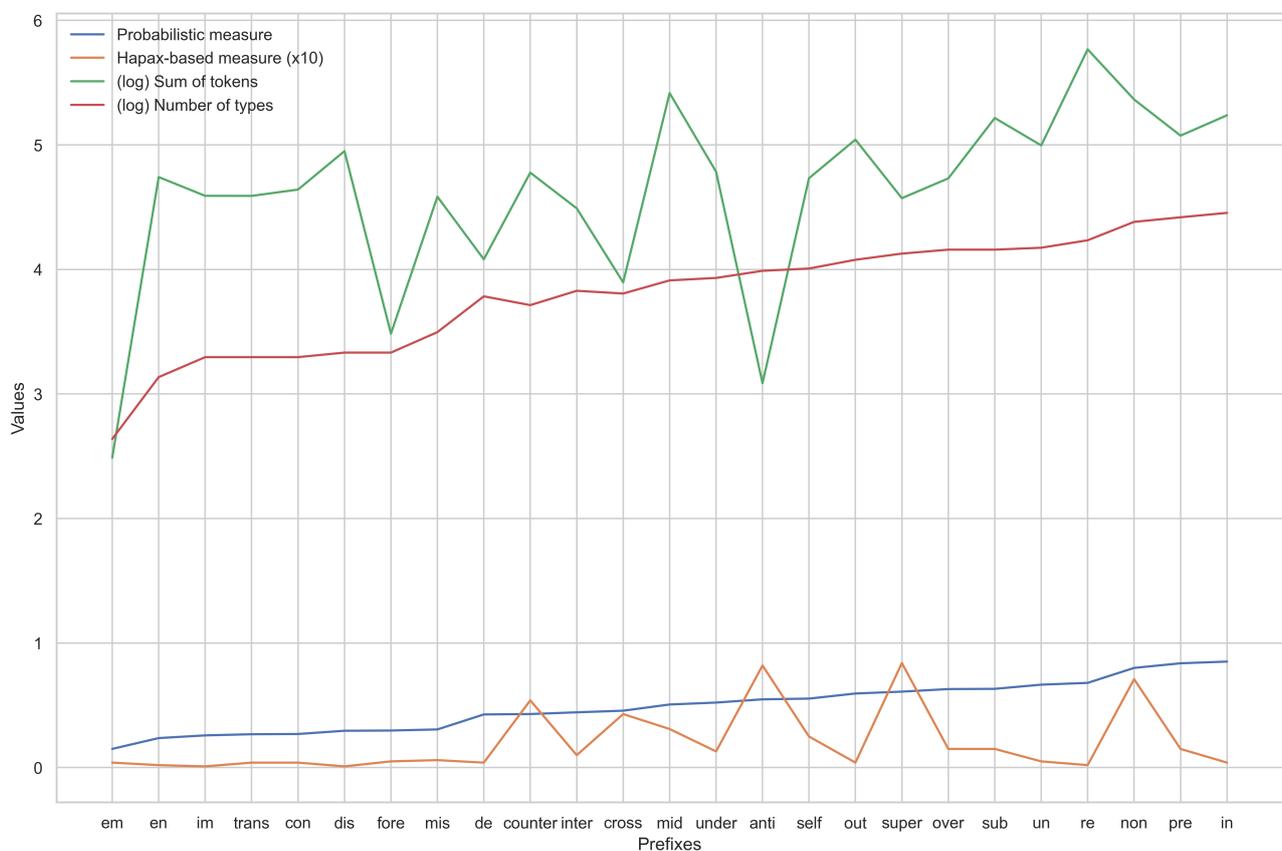

Figure 4. English prefixes' productivity measures plotted against the numbers of types and tokens

Hence, I cannot agree with the idea of Hay and Baayen (2002) that there exists a significant inverse relationship between the token frequency of an affix and this affix's morphological productivity: 'the more often you encounter an affix, <…> the less productive that affix is likely to be' (Hay and Baayen, 2002: 219). Probably, this inverse relationship is simply another manifestation of the fact that the hapax-based measure is biased by token frequency. My results bring me to a far less paradoxical conclusion: the more productive a prefix, the bigger its output, i.e. the greater number of types and the higher overall token frequency of the words with this prefix.

## 5 Conclusion

It is a long-established view that high token frequency represents a sort of stumbling block for affixes' linguistic productivity. It has been argued that affixes encountered in many frequent items become less parsable and, by that, lose their ability to combine with new bases, cf.: 'The less useful an affix is <…>, the more productive it is likely to be' (Hay and Baayen, 2002: 222). However, based on my findings, the picture appears to be more complicated: high-frequency derivations with an affix, once they are accumulated in a certain amount of types, do not block the emergence of new low-frequency coinages but rather facilitate them, serving as pathbreakers for neologisms.

In this paper, I suggest that linguistic productivity should be viewed as the probability of an affix to combine with a random base. Using the internet corpus of English from 2018, I evaluated the linguistic productivity of 25 English prefixes and 27 Russian prefixes. For each prefix, three probabilities were obtained: (1) $P(X = 0)$, the probability of no occurrence of the combination of this prefix with a random base in the corpus; (2) $P(X = 1)$, the probability that the combination of this prefix with a random base will be of low frequency; and (3) $P(X = 2)$, the probability that the combination of this prefix with a random base will be frequent.

The true measure of linguistic productivity was estimated in two steps. First, the initial and transition probability distributions for a two-time-slice dynamic Bayesian network were learned on a random sample of 100 random bases obtained from the corpus. Second, the value of $P(X = 1) + P(X = 2)$ was calculated for the 101st random base, given the last base in the sample. I found that, based on the evaluations of these probabilities, all prefixes, when arranged in order of ascending productivity, could be subdivided into three groups. The first group encompasses prefixes with the probabilities hierarchically arranged as $P(X = 0) > P(X = 2) > P(X = 1)$. In the second group, one finds prefixes where the probabilities are aligned in this way: $P(X = 2) > P(X = 0) > P(X = 1)$. Finally, the prefixes that belong to the last group reveal the following pattern: $P(X = 2) > P(X = 1) > P(X = 0)$.

Interestingly, these categorical differences were found to emerge as manifestations of an inherently gradient structure. Thus, within the first group, the differences between probabilities $P(X = 0)$ and $P(X = 2)$ continuously decrease, while the differences between probabilities $P(X = 2)$ and $P(X = 1)$ continuously increase. Within the second group, a similar mechanism of change can be observed, though with different contrasts. The differences between probabilities $P(X = 2)$ and $P(X = 0)$ become larger, while the differences between probabilities $P(X = 0)$ and $P(X = 1)$ become smaller. Finally, within the third group, the gap between probabilities $P(X = 2)$ and $P(X = 1)$ successively narrows, while the gap between probabilities $P(X = 1)$ and $P(X = 0)$ widens.

All of the above raises an interesting question of how derivational patterns spread. As Haspelmath noted, 'what is really remarkable about morphology is that morphological rules may <…> be unproductive' (Haspelmath, 2002: 40). As an example of an unproductive derivational rule in English, he mentioned the suffix -*al*, the list of action nouns formed with which is fixed and cannot be extended, so that no words like *repairal*, *ignoral*, and *amusal* are possible. No less remarkable, however, is the fact that the productivity of even fully productive affixes is not without its limits. Thus, it is not clear why, for example, given the high frequency of the verb *give* and the high productivity of the prefix *re*-, the derivation *re-give* is extremely unpopular, with zero hits per million tokens in both COCA and *ententen18_tt31*. It is also unclear why, given that the verb *evolve* is more frequent than the verb *regulate*, only *dis-regulate* is actually attested in COCA and *ententen18_tt31*, although there seems to be nothing conceptually improbable or semantically incompatible in the possible combination *dis-evolve*.

One might conclude that, even for very productive affixes, there is no simple linear relation between base and derivation frequency. Rather, it is high-frequency items with a certain affix that play a pivotal role in the self-propagating of respective derivational patterns and the structuring of its output, with less-frequent members being grouped around more prominent ones. I believe that this 'clustering hypothesis' deserves further investigation.